\documentclass[final]{elsarticle}

\usepackage{graphicx}
\usepackage{amsmath}
\usepackage{amssymb}
\usepackage{algorithmic}
\usepackage{algorithm}
\usepackage{subfigure}
\usepackage{array}
\usepackage{tabularx}
\usepackage{multirow}
\usepackage{pdfsync}
\usepackage[pagewise]{lineno}

\newcommand{\bfx}{{\textbf{x}}}
\newcommand{\bfv}{{\textbf{v}}}

\journal{Pattern Recognition}

\begin{document}

\begin{frontmatter}

\title{Max-Min Distance Nonnegative Matrix Factorization}

\author[Buffalo]{Jim Jing-Yan Wang}
\ead{jimjywang@gmail.com}

\address[Buffalo]{University at Buffalo, The State University of New York, Buffalo, NY 14203, USA}

\begin{abstract}
Nonnegative Matrix Factorization (NMF) has been a popular representation method for
pattern classification problem.
It tries to decompose a nonnegative matrix of data samples as the product of a nonnegative basic matrix and a nonnegative coefficient matrix, and
the coefficient matrix is used as the new representation.
However, traditional NMF methods ignore the class labels of the data samples.
In this paper, we proposed a supervised novel NMF algorithm to improve the
discriminative ability of the new representation.
Using the class labels, we separate all the data sample pairs into within-class pairs
and between-class pairs.
To improve the discriminate ability of the new NMF representations, we
hope that the maximum distance of the within-class pairs pairs in the
new NMF space could be minimized, while
the minimum distance of the between-class pairs pairs  could be maximized.
With this criteria, we construct an objective function and optimize it with regard to
basic and coefficient matrices, and slack variables alternatively, resulting in a iterative algorithm.
\end{abstract}


\end{frontmatter}

\linenumbers

\section{Introduction}

Nonnegative matrix factorization (NMF) \cite{lin2007projected,seung2001algorithms} has attracted much attention from
both research and engineering communities.
Given a data matrix with all its elements nonnegative, NMF tries to decompose it as a
product of two nonnegative low-rank matrices.
One matrix can be regarded as a basic matrix with its columns as basic vectors, and
the the other one as linear combination coefficient matrix so that the original data columns in the
original matrix could be represented as the linear contamination of the basic vectors.
Because of the nonnegative constrains on both the factorization metrics,
it only allow the additive linear combination, and thus a parts-based
representation could be archived \cite{cai2011graph}.
Since the original NMF approach was proposed by Seung and Lee \cite{lee1999learning,seung2001algorithms},
due to its ability to learn the parts of the data set \cite{li2001learning},
it has been used as a effective data representation method in various problems, such as
pattern recognition \cite{hoyer2004non},
computer vision \cite{shashua2005non}, bioinformatics \cite{gao2005improving}, etc.
The most popular application of NMF as a data representation
problem is pattern recognition, where the nonnegative feature vectors of the data samples are organized as
a nonnegative matrix,
and the columns of the coefficient matrix are used as the new low-dimension representations.

Among the pattern recognition problems, when NMF is performed to the data matrix,
it is usually assumed that the class labels of the data samples are not available,
making it an unsupervised problem.
Such typical application is clustering of images and documents \cite{cai2011graph,liu2012constrained}.
However, in real world supervised or semi-supervised classification applications, the  class labels
of the training data samples are usually available,
which is ignored by most existed NMF methods.
If the class label information could be unutilized during the
representation procedure, the
discriminative ability of the representation could be improved significantly.
To this end, some supervised and semi-supervised NMF are proposed.
For example, Wang and Jia \cite{wang2004fisher}
proposed the
Fisher nonnegative matrix factorization (FNMF) method to
encode
discrimination information for a classification problem
by imposing Fisher constraints on the NMF
algorithm.
Lee et al. \cite{lee2010semi} proposed the
semi-supervised nonnegative matrix factorization (SSNMF)
by jointly incorporating the data matrix and the (partial) class label matrix into NMF.
Most recently, Liu et al. \cite{liu2012constrained}
the  constrained nonnegative matrix factorization (CNMF) by incorporateing the label information as additional constraints.

In this paper, we propose to a novel supervised NMF method, by exploring the
class label information and using it to constrain the coefficient vectors of the data samples.
We consider the data sample pairs, and the class labels
of the samples allow us to  separate the pairs to two types --- the
within-class pair and the between-class pair.
To improve the discriminate ability of the coefficient vectors of the samples,
we consider the distance between the coefficient vectors of each sample pairs,
and try to minimize that of the the within-class pairs, while maximize the
that of the between-class pairs.
In this way, the coefficient vectors of data samples of the same class can be gathered, while that of different classes
can be separated.
One problem is how to assign different weights to differen pairs in the objective function.
To avoid this problem, we apply a strategy similar to max-min distance analysis \cite{bian2011max}.
The \textbf{maximum}
within-class pair coefficient vector distance is minimized,
so that all the within-class pair  coefficient vector distance can be minimized also.
Meanwhile
the \textbf{minimum}
between-class pair coefficient vector distance is maximized,
so that all the between-class pair  coefficient vector distance can be maximized also.
We construct a novel objective function for NMF to imposing both the maximum within-class pair distance minimization
and the minimum between-class pair distance maximization problems.
By optimizing it with an alternative strategy, we develop an iterative algorithm.
The proposed method is called Max-Min Distance NMF (MMDNMF).

The remaining parts of this paper is organized as follows:
In section \ref{sec:Met}, we introduce the novel NMF method. In section \ref{sec:Experiment},
the experimental results are given to verify the effectiveness of the problem method.
The paper is concluded in section \ref{sec:conclusion}.

\section{Max-Min Distance NMF}
\label{sec:Met}

In this section, we first formulate the problem with and objective function,
and then optimize it to obtain the iterative algorithm.

\subsection{Objective function}

Suppose we have $n$ data samples in the training set $\mathcal{X}=\{\bfx_i\}_{i=1}^n$,
where $\bfx_i\in \mathbb{R}^d_+$ is the $d$-dimensional nonnegative feature vector of
the $i$-th  sample,
we organize the samples as a nonnegative matrix $X=[\bfx_1,\cdots,\bfx_n]\in \mathbb{R}^{d\times n}_+$.
The $i$-th column of the matrix $X$ is the   feature vector of the $i$-th sample.
Their corresponding class label set is denoted as $\{y_i\}_{i=1}^n$,
where $y_i\in \mathcal{Y}$ is the class label of the $i$-th sample,
and $\mathcal{Y}$ is the class label space.
NMF aims to find two low rank nonnegative matrices $U\in \mathbb{R}^{d\times m}_+$ and $V \in \mathbb{R}^{m\times n}_+$,
where $m\leq d$,
so that the product of them, $UV$, could approximate the original matrix, $X$, as accurate as possible

\begin{equation}
\begin{aligned}
X\approx UV
\end{aligned}
\end{equation}
The $m$ columns of matrix $U$ could be regarded as $m$ basic vectors, and
each sample $\bfx_i$ could be represented as the nonnegative linear combination
of these basic vectors.
The linear combination coefficient vector of  $\bfx_i$ is the $i$-th
column vector $\bfv_i\in \mathbb{R}^m_+$ of $V$.
We can also regard $\bfv_i$ as the new low-dimensional presentation vector of $\bfx_i$ with regard to the basic matrix $U$.
To seek the optimal matrices $U$ and $V$,
we consider the following problems to construct our objective function:

\begin{itemize}
\item
To reduce  the approximation error between $X$ and $UV$,
the squared $L_2$ distance between them is usually
minimized with regard to $U$  and $V$ as follows,

\begin{equation}
\label{equ:NMF}
\begin{aligned}
\underset{U,V}{\min}
~&\|X- UV\|_2^2\\
s.t.
~&
U\geq 0,V\geq 0
\end{aligned}
\end{equation}

\item
We consider
the training sample pairs in the training set,
and
separate them to two pair sets --- the within-class pair set $\mathcal{W}$,
and the between-class pair set $\mathcal{B}$.
The within-class pair  set is defined as the set of sample pair belonging to the same class, i.e.,
$\mathcal{W}=\{(i,j)|y_i=y_j,\bfx_i,\bfx_j\in \mathcal{X}\}$.
The between-class pair  set is defined as the set of sample pairs belonging to different classes, i.e.,
$\mathcal{B}=\{(i,j)|y_i \neq y_j,\bfx_i,\bfx_j\in \mathcal{X}\}$.
To the two samples of the $(i,j)$-th pair in the new
coefficient vector space, we use the
squared $L_2$ norm distance between their coefficient vectors,
$\|\bfv_i-\bfv_j\|_2^2$.
Apparently, to improve the discriminate ability  of the
new NMF presentation, the  coefficient vector distance of within-class pairs should be
minimized while that of the between-class pairs
should be maximized.
Instead of considering all the pairs, we
directly
minimize the maximum   coefficient vector distance of within-class
pairs, and thus
we duly considers the
aggregation of all within-class pairs, as follows,

\begin{equation}
\label{equ:Within}
\begin{aligned}
\underset{V}{\min}
~&
\left \{
\max_{(i,j)\in \mathcal{W}} \|\bfv_i-\bfv_j\|^2_2
\right \}\\
s.t.
~&
V\geq 0
\end{aligned}
\end{equation}
Meanwhile,  we also
maximize the minimum   coefficient vector distance of between-class
pairs, and thus
we  consider the
separation of all between-class pairs, as follows,

\begin{equation}
\label{equ:Between}
\begin{aligned}
\underset{V}{\max}
~&\left \{
\min_{(i,j)\in \mathcal{B}} \|\bfv_i-\bfv_j\|^2_2
\right \}\\
s.t.
~&
V\geq 0
\end{aligned}
\end{equation}
In this way, the maximum within-class pair distance is minimized, so that all
the within-class pair distances are also minimized.
Similarly,
the minimum between-class pair distance is maximized, so that all
the between-class pair distances are also minimized.

\end{itemize}

To formulate our problem, we combine the problems in (\ref{equ:NMF}),
(\ref{equ:Within}) and (\ref{equ:Between}), and propose
the novel optimization problem for NMF as

\begin{equation}
\label{equ:objective}
\begin{aligned}
\underset{U,V}{\min}
~&\left \{
\|X- UV\|_2^2
+a \max_{(i,j)\in \mathcal{W}} \|\bfv_i-\bfv_j\|^2_2
-b \min_{(i,j)\in \mathcal{B}} \|\bfv_i-\bfv_j\|^2_2
\right\}\\
s.t
~&
U\geq 0,V\geq 0
\end{aligned}
\end{equation}
where $a$ and $b$ are the trade-off parameters.
It should be noted that in (\ref{equ:objective}), the maximization and
minimization problem are coupled, making it difficult to optimize.
To solve this problem, we introduce to nonnegative slake variables $\varepsilon\geq 0$
and $\zeta \geq 0$ to represent the maximum coefficient vector distance between
all within-class pairs,
and the minimum  coefficient vector distance between
all between-class pairs.
In this way, (\ref{equ:objective}) could be rewritten as

\begin{equation}
\label{equ:objective1}
\begin{aligned}
\underset{U,V,\varepsilon, \zeta}{\min}
~&\left \{
\|X- UV\|_2^2
+a\varepsilon
-b\zeta
\right\}\\
s.t.
~&
\|\bfv_i-\bfv_j\|^2_2\leq \varepsilon, \forall (i,j)\in \mathcal{W},\\
&\|\bfv_i-\bfv_j\|^2_2 \geq \zeta, \forall (i,j)\in \mathcal{B},\\
&
U\geq 0,V\geq 0,\varepsilon\geq 0, \zeta \geq 0.
\end{aligned}
\end{equation}
In this problem, the two slake variables are also optimized
with the
basic matrix $U$ and the coefficient matrix $V$.

\subsection{Optimization}

To solve the problem introduce in (\ref{equ:objective1}), we come up with
the Lagrange function as follows,

\begin{equation}
\label{equ:Lagrange}
\begin{aligned}
\mathcal{L}&
\left (U,V,\varepsilon,\zeta,\lambda_{ij},
\xi_{ij},\Sigma,\Upsilon,\phi,\varphi
\right )\\
=
&
\|X- UV\|_2^2
+a\varepsilon
-b\zeta
\\
&+\sum_{(i,j)\in \mathcal{W}} \lambda_{ij}
\left (\|\bfv_i-\bfv_j\|^2_2- \varepsilon \right )
-
\sum_{(i,j)\in \mathcal{B}} \xi_{ij}
\left ( \|\bfv_i-\bfv_j\|^2_2 - \zeta \right )\\
&-Tr(\Sigma U^\top)-Tr(\Upsilon V^\top)\\
&-\phi \varepsilon - \varphi \zeta
\end{aligned}
\end{equation}
where $\lambda_{ij}\geq 0$ is the Lagrange multiplier for the
constrain $\|\bfv_i-\bfv_j\|^2_2\leq \varepsilon$,
$\xi_{ij}\geq 0$ is the Lagrange multiplier for the
constrain $ \|\bfv_i-\bfv_j\|^2_2 \geq \zeta$,
$\Sigma\in \mathbb{R}^{d\times m}_+$ is the Lagrange multiplier matrix for $U\geq 0$,
$\Upsilon\in \mathbb{R}^{m\times n}_+$ is the Lagrange multiplier matrix for $V\geq 0$,
$\phi\geq 0$  is the Lagrange multiplier  for $\varepsilon \geq 0$,
and $\varphi\geq 0$ is the Lagrange multiplier  for $\zeta \geq 0$.
According to the duality theory of optimization \cite{diewert1974applications}, the optimal solution could be
achieved by solving the following problem,

\begin{equation}
\label{equ:optim}
\begin{aligned}
\underset{
\left.\begin{matrix}
\lambda_{ij},
\xi_{ij},
\\
\Sigma,\Upsilon,\phi,\varphi
\end{matrix}\right.
}{\max}
\underset{U,V,\varepsilon,\zeta}{\min}
~
&
\mathcal{L}
\left (U,V,\varepsilon,\zeta,\lambda_{ij},
\xi_{ij},\Sigma,\Upsilon,\phi,\varphi
\right )\\
s.t.~
&
\lambda_{ij}\geq 0, \forall (i,j)\in \mathcal{W},\\
&
\xi_{ij}\geq 0, \forall(i,j)\in \mathcal{B},\\
&\Sigma\geq 0, \Upsilon\geq 0,\phi\geq 0,\varphi\geq 0.
\end{aligned}
\end{equation}
By substituting (\ref{equ:Lagrange}) to (\ref{equ:optim}), we obtain the following problem,

\begin{equation}
\label{equ:objective2}
\begin{aligned}
\underset{
\left.\begin{matrix}
\lambda_{ij},
\xi_{ij},
\\
\Sigma,\Upsilon,\phi,\varphi
\end{matrix}\right.
}{\max}
\underset{U,V,\varepsilon,\zeta}{\min}
~
&
\left \{
\|X- UV\|_2^2
+a\varepsilon
-b\zeta
\vphantom{
\sum_{(i,j)\in \mathcal{W}} \lambda_{ij}
\left (\|\bfv_i-\bfv_j\|^2_2- \varepsilon \right )
}
\right.
\\
&+\sum_{(i,j)\in \mathcal{W}} \lambda_{ij}
\left (\|\bfv_i-\bfv_j\|^2_2- \varepsilon \right )
-
\sum_{(i,j)\in \mathcal{B}} \xi_{ij}
\left ( \|\bfv_i-\bfv_j\|^2_2 - \zeta \right )\\
&
\left.
-Tr(\Sigma U^\top)-Tr(\Upsilon V^\top)
-\phi \varepsilon - \varphi \zeta
\vphantom{
\sum_{(i,j)\in \mathcal{W}} \lambda_{ij}
\left (\|\bfv_i-\bfv_j\|^2_2- \varepsilon \right )
}
\right \}
\\
s.t.~
&
\lambda_{ij}\geq 0, \forall (i,j)\in \mathcal{W},\\
&
\xi_{ij}\geq 0, \forall(i,j)\in \mathcal{B},\\
&\Sigma\geq 0, \Upsilon\geq 0,\phi\geq 0,\varphi\geq 0.
\end{aligned}
\end{equation}
This problem is difficult to optimize directly. Instead of solving it
with regard to all the variables simultaneously, we adopt an alternate optimization
strategy \cite{lootsma1994alternative}.
The NMF factorization matrices $U$ and $V$, slack variables $\phi$ and $\varphi$,
and the Lagrange  multipliers  $\lambda_{ij}$ and $\xi_{ij}$
are updated alternatively in an iterative algorithm.
When one variable is optimized, other variables are fixed.

\subsubsection{Optimizing $U$ and $V$}

By fixing other variables and removing the terms irrelevant to $U$ or $V$, the optimization problem in
(\ref{equ:objective2}) is reduced to

\begin{equation}
\label{equ:objectiveUV}
\begin{aligned}
\underset{
\Sigma,\Upsilon
}{\max}
~
\underset{U,V}{\min}
~
&
\left \{
\|X- UV\|_2^2
+\sum_{(i,j)\in \mathcal{W}} \lambda_{ij}
\|\bfv_i-\bfv_j\|^2_2
-
\sum_{(i,j)\in \mathcal{B}} \xi_{ij}
\|\bfv_i-\bfv_j\|^2_2
\right.\\
&
~~~-Tr(\Sigma U^\top)-Tr(\Upsilon V^\top)\\
&
=Tr(X X^\top) - 2 Tr(X V^\top U^\top) + Tr (UV V^\top U^\top)
\\
&
\left.
~~~
+2 Tr\left ( V (D-\Lambda) V^\top \right )
-2 Tr\left ( V (E-\Xi) V^\top \right )
-Tr(\Sigma U^\top)-Tr(\Upsilon V^\top)
\vphantom{
\sum_{(i,j)\in \mathcal{W}} \lambda_{ij}
\left (\|\bfv_i-\bfv_j\|^2_2- \varepsilon \right )
}
\right \}
\\
s.t.~
&\Sigma\geq 0, \Upsilon\geq 0.
\end{aligned}
\end{equation}
where $\Lambda\in \mathbb{R}^{n\times n}_+$ and $\Xi\in \mathbb{R}^{n\times n}_+$ with

\begin{equation}
\begin{aligned}
\Lambda_{ij}=
\left\{\begin{matrix}
\lambda_{ij}, &if~(i,j)\in \mathcal{W} \\
0, &otherwise
\end{matrix}\right.
,~
\Xi_{ij}=
\left\{\begin{matrix}
\xi_{ij}, &if~(i,j)\in \mathcal{B} \\
0, &otherwise
\end{matrix}\right.
\end{aligned}
\end{equation}
$D \in \mathbb{R}^{n\times n}_+$ is a
diagonal matrix whose entries are column  sums of $\Lambda$,
 $D_{ii} = \sum_{i}\Lambda_{ij} $,
 and
$E \in \mathbb{R}^{n\times n}_+$ is a
diagonal matrix whose entries are column  sums of $\Xi$,
$E_{ii} = \sum_{i}\Xi_{ij} $.
To solve this problem, we set the partial derivatives of
the objective function in (\ref{equ:objectiveUV})
 with respect to $U$ and $V$ to zero, and we have

\begin{equation}
\begin{aligned}
&-2 X V^\top + 2 UV V^\top - \Sigma = 0\\
&
- 2 U^\top X  + 2 U^\top UV
+2  V (D-\Lambda) - 2  V (E-\Xi)  - \Upsilon = 0\\
\end{aligned}
\end{equation}
Using the KKT conditions $  [\Sigma] \circ [U ] =$ and $  [\Upsilon] \circ [V] = 0$  \cite{bach2004multiple},
where $[~]\circ [~]$ denotes the element-wise product between two matrices, we get
the following equations for $U$ and $V$:

\begin{equation}
\begin{aligned}
&- [X V^\top] \circ [U ] + [ UV V^\top] \circ [U ] = 0\\
&
-  [U^\top X] \circ [V]  +  [U^\top UV] \circ [V]
+[  V (D-\Lambda)]\circ [V] - [  V (E-\Xi)] \circ [V]  = 0
\end{aligned}
\end{equation}
which lead to the following updating rules:

\begin{equation}
\begin{aligned}
&
U  \leftarrow   \frac{[X V^\top]}{[UV V^\top]}\circ [U]\\
&
V  \leftarrow   \frac{[U^\top X +   V \Lambda + V E ]}{[U^\top UV +V D + V \Xi]}\circ [V]
\end{aligned}
\end{equation}
where $\frac{[~]}{[~]}$ is the element-wise matrix division operator.

\subsubsection{Optimizing $\varepsilon$ and $\zeta$}

By removing terms irrelevant to $\varepsilon$ and $\zeta$ and fixing all other variables,
we have the following optimization problem with regard to only  $\varepsilon$ and $\zeta$:

\begin{equation}
\label{equ:var}
\begin{aligned}
\underset{\phi,\varphi}{\max}
~
\underset{\varepsilon,\zeta}{\min}
~&
\left \{
a\varepsilon
-b\zeta
-\sum_{(i,j)\in \mathcal{W}} \lambda_{ij}
\varepsilon +
\sum_{(i,j)\in \mathcal{B}} \xi_{ij}
\zeta
-\phi \varepsilon - \varphi \zeta
\right \}\\
s.t.~
&\phi\geq 0,\varphi\geq 0.
\end{aligned}
\end{equation}
By setting the partial derivatives of
the objective function in (\ref{equ:var})
 with respect to $\varepsilon$ and $\zeta$ to zero, we have

\begin{equation}
\label{equ:constrain}
\begin{aligned}
&
a
-\sum_{(i,j)\in \mathcal{W}} \lambda_{ij}
-\phi=0\\
&-b
+\sum_{(i,j)\in \mathcal{B}} \xi_{ij}
-\varphi=0
\end{aligned}
\end{equation}
Using the KKT conditions $\phi \varepsilon = 0$ and $\varphi \zeta = 0$,
we get
the following equations for $\varepsilon$ and $\zeta$:

\begin{equation}
\begin{aligned}
&a\varepsilon -
\left( \sum_{(i,j)\in \mathcal{W}} \lambda_{ij}\right ) \varepsilon = 0\\
&
-
b\zeta
 +
 \left(  \sum_{(i,j)\in \mathcal{B}} \xi_{ij} \right )  \zeta = 0
\end{aligned}
\end{equation}
which lead to the following updating rules:

\begin{equation}
\begin{aligned}
&
\varepsilon\leftarrow
\frac{
\left( \sum_{(i,j)\in \mathcal{W}} \lambda_{ij}\right )}
{a} \varepsilon\\
&
\zeta\leftarrow
\frac{b}{
\left (\sum_{(i,j)\in \mathcal{B}} \xi_{ij}\right )} \zeta\\
\end{aligned}
\end{equation}

\subsubsection{Optimizing $\lambda_{ij}$ and
$\xi_{ij}$}

Based on (\ref{equ:constrain}), we have the following constrains for $\lambda_{ij}$ and
$\xi_{ij}$,

\begin{equation}
\begin{aligned}
&
\phi = a - \sum_{(i,j)\in \mathcal{W}} \lambda_{ij}  \geq 0
\Rightarrow \sum_{(i,j)\in \mathcal{W}} \lambda_{ij} \leq a,\\
&
\varphi = -b + \sum_{(i,j)\in \mathcal{B}} \xi_{ij}  \geq 0
\Rightarrow  \sum_{(i,j)\in \mathcal{B}} \xi_{ij}  \geq b.
\end{aligned}
\end{equation}
By
Considering these constrains, fixing other variables and removing terms irrelevant to $\lambda_{ij}$ and
$\xi_{ij}$ from (\ref{equ:optim}), we have the following problem with regard to $\lambda_{ij}$ and
$\xi_{ij}$,

\begin{equation}
\begin{aligned}
\underset{
\lambda_{ij},
\xi_{ij}
}{\max}
&
\left \{
\sum_{(i,j)\in \mathcal{W}} \lambda_{ij}
\left(\|\bfv_i-\bfv_j\|^2_2-\varepsilon\right)-
 \sum_{(i,j)\in \mathcal{B}} \xi_{ij}
\left ( \|\bfv_i-\bfv_j\|^2_2 - \zeta\right)
\right \}\\
s.t.
&\lambda_{ij}\geq 0, \forall (i,j)\in \mathcal{W},
~
\xi_{ij}\geq 0, \forall (i,j)\in \mathcal{B},\\
&
\sum_{(i,j)\in \mathcal{W}} \lambda_{ij} \leq a,
\sum_{(i,j)\in \mathcal{B}} \xi_{ij}  \geq b.
\end{aligned}
\end{equation}
This problem can be solved as a linear programming (LP) problem.

\section{Conclusion}
\label{sec:conclusion}

In this paper, we investigate how to use the class labels of the data samples
to improve the discriminative ability of their  NMF representations.
To explore the class label information of the data samples, we consider the within-class sample
pairs with the same class labels, and
also the between-class sample pairs with different class labels.
Apparently, in the NMF representation space, we need to minimize the distances between the   within-class pairs,
and also maximize the distances between
the between-class pairs.
Inspired by the max-min distance analysis \cite{bian2011max}, we also consider the
extreme situation: we pick up the maximum within-class distance and then try to minimize it,
so that all the within-class distances are also minimized,
and we pick up the minimum between-class distance and then maximize it,
so that all the between-class distances are maximized.
Differently from the max-min distance analysis, which only pick up the maximize the between-class
distance and minimize it, we consider the between-class and within class distances dually.

\end{document}